\documentclass[10pt,letterpaper]{article}

\usepackage{cogsci}

\cogscifinalcopy %

\usepackage{pslatex}
\usepackage{apacite}
\usepackage{float} %

\usepackage[round]{natbib}
\usepackage{xcolor}
\usepackage{amsmath}
\usepackage[acronym]{glossaries}
\usepackage{array}
\usepackage{booktabs}
\usepackage[hidelinks,breaklinks]{hyperref}

\usepackage{amsmath,amsfonts,bm}

\def\eqref#1{equation~\ref{#1}}

\def\1{\bm{1}}

\def\rvtheta{{\boldsymbol{\theta}}}
\def\rvphi{{\boldsymbol{\phi}}}

\def\rvg{{\mathbf{g}}}

\def\rvs{{\mathbf{s}}}

\def\rmA{{\mathbf{A}}}

\DeclareMathAlphabet{\mathsfit}{\encodingdefault}{\sfdefault}{m}{sl}
\SetMathAlphabet{\mathsfit}{bold}{\encodingdefault}{\sfdefault}{bx}{n}

\newacronym{sdt}{SDT}{Self-Determination Theory}
\newacronym{im}{IM}{intrinsic motivation}
\newacronym{ai}{AI}{artificial intelligence}
\newacronym{rl}{RL}{reinforcement learning}
\newacronym{gcrl}{GCRL}{goal-conditioned reinforcement learning}
\newacronym{ride}{RIDE}{Rewarding Impact-Driven Exploration}
\newacronym{lp}{LP}{Learning Progress}
\newacronym{vic}{VIC}{Variational Intrinsic Control}
\newacronym{diayn}{DIAYN}{Diversity Is All You Need}
\newacronym{cbim}{CB-IM}{competence-based intrinsic motivation}
\newacronym{rig}{RIG}{Reinforcement Learning With Imagined Goals}
\newacronym{curious}{CURIOUS}{Continual Universal Reinforcement learning with Intrinsically mOtivated sUbstitutionS}
\newacronym{acl}{ACL}{Automatic Curriculum Learning}
\newacronym{ued}{UED}{Unsupervised Environment Design}
\newacronym{omni-epic}{OMNI-EPIC}{Open-endedness via Models of human Notions of Interestingness with Environments Programmed in Code}
\newacronym{imrl}{IMRL}{Intrinsically Motivated Reinforcement Learning}
\newacronym{cpm}{CPM}{Competence Progress Motivation}

\title{Towards a Formal Theory of the Need for Competence\\
via Computational Intrinsic Motivation}
 
\author{Erik M.~Lintunen\textsuperscript{1} \and Nadia M.~Ady\textsuperscript{1} \and Sebastian Deterding\textsuperscript{2} \and Christian Guckelsberger\textsuperscript{1}\\
\textsuperscript{1}Department of Computer Science, Aalto University, Finland, {\{firstname.lastname\}}@aalto.fi\\
\textsuperscript{2}Dyson School of Design Engineering, Imperial College London, United Kingdom, s.deterding@imperial.ac.uk
}

\begin{document}

\maketitle

\begin{abstract}
Computational modelling offers a powerful tool for formalising psychological theories, making them more transparent, testable, and applicable in digital contexts.
Yet, the question often remains: how should one computationally model a theory?
We provide a demonstration of how formalisms taken from artificial intelligence can offer a fertile starting point.
Specifically, we focus on the ``need for competence'', postulated as a key basic psychological need within Self-Determination Theory (SDT)---arguably the most influential framework for intrinsic motivation (IM) in psychology.
Recent research has identified multiple distinct facets of competence in key SDT texts: effectance, skill use, task performance, and capacity growth.
We draw on the computational IM literature in reinforcement learning to suggest that different existing formalisms may be appropriate for modelling these different facets.
Using these formalisms, we reveal underlying preconditions that SDT fails to make explicit, demonstrating how computational models can improve our understanding of IM.
More generally, our work can support a cycle of theory development by inspiring new computational models, which can then be tested empirically to refine the theory.
Thus, we provide a foundation for advancing competence-related theory in SDT and motivational psychology more broadly.

\end{abstract}

\section{Introduction}

\Gls{sdt} \citep{deci1985intrinsic} is one of the most influential psychological theories of motivation, specifically \gls{im}. Notably, \gls{sdt}, like many psychological theories, is expressed in ``soft'' verbal propositions. In light of the current theory crisis \citep{oberauer2019addressing}, only-verbal theory has drawn increasing critique: it can leave many auxiliary assumptions and boundary conditions unspecified, making it hard to derive hypotheses that can be rigorously tested \citep{scheel2021hypothesis}. Many psychology researchers have advocated for computational modelling as a paradigm (e.g., \citealp{marsella2010computational,oberauer2019addressing,robinaugh2021invisible}) or even as a necessity \citep{van2024productive} for theory development. Yet formal specification as computational modelling remains largely absent in \gls{sdt}, judging by, e.g., the most recent integrative volume of the theory \citep{ryan2023oxford}.

A major challenge of computational modelling is inferring plausible formalisms for a given theory.
We make the case that computational \gls{im}, i.e., formalisms developed to drive the behaviour of artificial agents, can inform the development of \gls{sdt} theory in human motivation research. Computational \gls{im} has been informed by \gls{sdt} (e.g., \citealp{oudeyer2007what}), but has undergone major advancements independent from human psychology, addressing problems such as open-ended development, task-agnostic learning, and dealing with the potential sparsity of rewards \citep[p.~1161]{colas2022autotelic}.

Our case study focuses on the \emph{need for competence}, one of the basic psychological needs underpinning \gls{im} in \gls{sdt}. Through conceptual analysis \citep{machado2007toward}, \citet{deterding2024why} identify four distinct facets of competence in canonic \gls{sdt} texts, reproduced below in \autoref{table:facets}.

\begin{table}[h]
\centering
\caption{\textbf{The four distinct facets of competence} identified by \citet[pp.~7--10]{deterding2024why}.} 
\label{table:facets} 
\vskip 0.12in
\begin{tabular}{@{}>{\raggedright\arraybackslash}p{0.41\linewidth}>{\raggedright\arraybackslash}p{0.53\linewidth}@{}}
\toprule
\textbf{Effectance (C1)} & Observing that one's action causes consequences, which can be unintended. \\
\midrule
\textbf{Skill use (C2)} & Observing \textbf{(a)} an opportunity (an appropriate situation) to use a skill and/or \textbf{(b)} that one uses a particular capacity in the course of one's action. \\
\midrule
\textbf{Task performance (C3)} & Observing that one performs \textbf{(a)} well at an intended task and/or \textbf{(b)} to an extent that requires a certain skill or skill level. \\
\midrule
\textbf{Capacity growth (C4)} & Observing gain in the \textbf{(a)} strength and/or \textbf{(b)} range of one's skills. \\
\bottomrule
\end{tabular}
\end{table}

In the section \textit{\nameref{sec:alignmodelswithsdt}}, we match each distinct facet with examples of formalisms from the computational \gls{im} literature and discuss how each example formalises aspects of the theory. These formalisms not only lay a foundation for computationally specifying the need for competence, but, by linking to the corresponding papers, we also unveil a wealth of existing implementations, simulators, and evaluation metrics that could be used in psychological motivation research. The value of such resources is detailed in the section \textit{\nameref{sec:futureconclusion}}, but, for one, they empower future testing of whether formalisms fitting the facets produce empirical data aligned with propositions from the theory.

\section{Background: Computational IM \& RL}
\label{sec:imgcrl}

We begin by justifying why \gls{rl} is a suitable domain from which to draw computational models of competence. In addition, to support understanding of the models we discuss, we introduce key concepts from intrinsically motivated \gls{rl}---notably, \emph{goals} and \emph{skills} and their relationship to the reward function---and point to further reading.

The idea that \gls{im} might be driven by biological reward systems (\citealp[p.~113]{barto2004intrinsically}, in reference to \citealp{kakade2002dopamine,dayan2002reward})
has driven the choice of methods centrally used in computational \gls{im}: \gls{rl} \citep[p.~252]{baldassarre2022intrinsic}. Due to their usefulness, intrinsically motivated \gls{rl} methods have been studied extensively in the computational \gls{rl} literature at large (see reviews by \citealp{colas2022autotelic,aubret2023information,lidayan2024bamdp}). Central to \gls{rl} is the construct of \textit{reward function}, and an assumption underlying much work in \gls{rl} is the \emph{reward hypothesis}: ``all of what we mean by goals and purposes can well be thought of as the maximization of the expected value of the cumulative sum of a received scalar signal (called reward)'' (\citealp[p.~53]{sutton2018reinforcement}; cf., \citealp[p.~4]{silver2021reward}). This hypothesis informs the formalisms used for \gls{cbim}, which tend to rely on the assumption that every goal can be sufficiently defined by a reward function. \emph{Intrinsically motivated} \gls{rl} is characterised by the use of task-agnostic intrinsic reward functions that collate or compare agent-internal variables independently of their semantics (\citealp[p.~3]{oudeyer2008how}; cf., \citealp[p.~246]{berlyne1965structure} and \citealp{oudeyer2007what}).

Many of our examples operate within the framework of \gls{gcrl} (see review by \citealp{liu2022goal}), which extends the standard definition of a reward function to be conditioned on goals. A \textit{goal}, or a \textit{goal-defining variable}, is a parameter to the reward function (cf., \citealp[p.~1165]{colas2022autotelic}; \citealp[p.~6]{aubret2023information}). Then, a \textit{skill} is a policy given a goal, optimising for the return according to the reward conditioned on that goal. Examples of goal-defining variables include indices \citep[e.g.,][]{eysenbach2019diversity} or elements drawn from a learned distribution \citep[e.g.,][]{nair2018visual}. Their role is simply to indicate which reward function the agent is aiming to maximise.

Goals can be viewed as ``a set of \emph{constraints} ... that the agent seeks to respect'' \citep[p.~1165, emphasis in original]{colas2022autotelic}. While the most immediate intuition of a goal is often as a desired state for the agent to reach (e.g., \citealp[p.~1]{kaelbling1993learning}; \citealp[p.~2]{schaul2015universal}), the formalism allows for a more general set of constraints on behaviour. In effect, any behaviour that can be defined by attempting to maximise some reward function on the environment can be formulated as a goal--skill pairing.

\section{Matching Computational Models With SDT}
\label{sec:alignmodelswithsdt}

We next turn our focus to matching computational formalisms to each of the competence facets in \autoref{table:facets} by providing examples from \gls{rl}. The formalisms have been identified through review of both classic \citep{oudeyer2007what} and state-of-the-art \citep{colas2022autotelic,aubret2023information} computational \gls{im}. An exhaustive matching is beyond the scope of this paper. Rather, we highlight formalisms that we believe to fit the verbal definitions of each facet and reveal preconditions that \gls{sdt} implicitly assumes. We focus on intuition about how each model fits the theory. For more intuition and detail on how each model is designed, implemented, and evaluated, we direct the reader to the corresponding papers.

\subsection{Effectance (C1)}
\label{sec:effectance}

Let us consider the case where competence in \gls{sdt} refers to \emph{effectance}, i.e., motivation by observing that one's own actions cause change in the environment. To recognise effectance, an agent needs a mechanism to distinguish whether a change in the environment is caused by the agent itself or some external factor. An example of a computational model fitting effectance is \gls{ride} \citep{raileanu2020ride}, as part of which the agent is rewarded for the amount of change it causes via a single action: the ``impact'' of the agent's action, measured between two consecutive observations. Specifically, the \emph{impact-driven reward} is defined as:
\begin{align}\label{eq:ride}
R_t(\rvs_t, \rvs_{t+1}) := \frac{\left\lVert\phi(\rvs_{t+1})-\phi(\rvs_t)\right\rVert_2}{\sqrt{N(\rvs_{t+1})}},
\end{align}
where $\phi(\rvs_t)$ and $\phi(\rvs_{t+1})$ are the learned representations of two consecutive states and $N(\rvs_{t+1})$ represents or approximates the number (count) of times the latter state has been visited \citep[pp.~4--5]{raileanu2020ride}.

In \gls{ride}, the agent computes its reward using the learned representation of the observation ($\phi(\rvs_t)$) as opposed to the raw sensory space (e.g., pixels). The parameters of the representation model $\phi$ are optimised so that an agent only learns the features it can control---that is, it learns to ignore aspects of the environment it cannot affect (inspired by \citealp{pathak2017curiosity}). The agent is rewarded for a combination of changes in only those state features that its own actions are able to consistently have some effect on (Eq.~\ref{eq:ride}, numerator). This reward is scaled according to the novelty of the state into which its action leads (Eq.~\ref{eq:ride}, denominator). The effect of the denominator is that the reward obtainable from particular states diminishes the more the agent observes them, even if the agent's action has the same effective outcome as previously observed. In recent \gls{sdt} texts, effectance motivation does not require novelty, but the original usage of the term required observations to be novel \citep[p.~322]{white1959motivation}. By having the reward wane the more a particular situation is perceived, \gls{ride} offers a simple formalisation of this original novelty aspect of effectance motivation, while ensuring the agent is only rewarded for changes its own actions have caused.

\subsection{Skill use (C2)}
\label{sec:skilluse}

We next consider the case in which competence refers to \emph{skill use} or the ``exercise $\ldots$ of one's capacities'' \citep[p.~86]{ryan2017self}. Following \citet[p.~6]{deterding2024why}, we cover the specific understanding of ``exercise'' as implementing or bringing something to bear \citep{webster2025}.\footnote{An alternative interpretation of ``exercise'', as engaging a skill repeatedly to grow it, will be treated under \textit{\nameref{sec:capacitygrowth}}.} As per \autoref{table:facets}, competence as skill-use motivation can be understood in two different ways: observing an opportunity (an appropriate situation) to use a skill (\textit{\hyperref[sec:c2a]{C2.a}}) or that one uses a particular capacity in the course of one's action (\textit{\hyperref[sec:c2b]{C2.b}}).

\subsubsection{An opportunity (an appropriate situation) to use a skill (C2.a)}
\phantomsection \label{sec:c2a}

Broadly, there are two kinds of computational approaches used to model an agent learning about opportunities to use particular skills: those in which the agent ignores the current state of the world and those in which it does not. In both cases the agent learns a policy over its set of skills to determine which skill to use. In the former case, an agent follows the same policy regardless of its state observation---in this case, an ``appropriate'' opportunity can be thought of as a learning opportunity, determined not based on the observation of the environment, but on how the agent's learning is proceeding. In the latter case, identifying an opportunity to use a particular skill can also depend on the observation.

Specifically, the first case of \textit{\hyperref[sec:c2a]{C2.a}}, in which an agent is learning about opportunities to use a skill while ignoring the current state of the world, is well exemplified by computational approaches using \gls{lp}, which approximates some derivative of performance over time (initially proposed by \citealp[p.~7]{oudeyer2007intrinsic}; see also related earlier work by \citealp[p.~1461]{schmidhuber1991curious}). In systems employing \gls{lp}, it is typically used as a measure to support selecting intermediate-difficulty goals with respect to the changing skills of an agent over time.
In fact, \gls{lp}, as an intrinsic reward for goal selection, has been well studied across both machine and human reinforcement learners (e.g., \citealp{colas2019curious,ten2021humans,molinaro2024latent}), and could thus offer good starting points for formalising \textit{\hyperref[sec:c2a]{C2.a}}.

While a (high-level) reward function prioritises the choice of skills, an agent also needs a mechanism to select a skill from that prioritisation. One commonly used mechanism is Thompson sampling \citep{thompson1933on}, also known as probability matching, in which an agent 
draws a skill according to a ``belief''---a probability distribution (learned from a reward signal such as \gls{lp} values) over its set of skills---and then acts greedily with respect to the drawn skill until it has more ``evidence'' (e.g., a change in \gls{lp} values) and its ``belief'' (i.e., skill-selection policy) is updated. Thompson sampling has been shown to explain aspects of human exploration in certain tasks \citep{gershman2018deconstructing}, making it a candidate for a psychologically plausible formalism of human skill selection.

An example of a system formalising the second case of \textit{\hyperref[sec:c2a]{C2.a}}, in which an agent learns about opportunities to use particular skills while accounting for the current state of the world, is \gls{vic}. In \gls{vic}, the appropriate skill for a given starting state is decided using a separate learned distribution dependant on how empowering the skill has been in the past. We mean \emph{empowering} in the sense of its definition in the \gls{ai} literature,\footnote{``Empowerment'' has also been studied in the context of human exploration, e.g., as ``exploring options that enable the generation of as many more options as possible'' \citep[p.~1482]{brandle2023empowerment}.} as the combined ability of an agent to control its environment and to sense this control (cf., \citealp[p.~2]{salge2014empowerment}; \citealp[p.~2]{gregor2016variational}). The decision over skills is conditional on the state of the world, in that the policy for choosing skills is updated based on how empowering the skill turned out to be \emph{for that given situation}. The distribution over skills is updated after skill use, such that more empowering skills become more likely to be chosen \citep[p.~4]{gregor2016variational}, using the reward \citep[p.~2]{gregor2016variational}:
\begin{align}\label{eq:vicreward}
R_t(\rvs_0,\rvs_f,\rvg) := \underbrace{\log q_\rvphi(\rvg \mid \rvs_0,\rvs_f)}_\text{($\alpha$)} \underbrace{-\log p_\rvtheta(\rvg \mid \rvs_0)}_\text{($\beta$)}.
\end{align}
\begin{description}
  \item[($\alpha$)] The discriminator, $q_\rvphi$, is an arbitrary variational distribution parametrised by $\rvphi$ \citep[p.~2]{barber2003im}. Given the first ($\rvs_0$) and last ($\rvs_f$) observations induced by the skill corresponding to the goal, $\rvg$, being pursued, $q_\rvphi$ defines a probability distribution over goals. Discriminating goals in the observation space requires the agent to observe distinct regions of the observation space. If a goal is not discriminable based on observations, two or more skills are producing overlapping behaviours; conversely, if a goal is discriminable, then the corresponding skill is inducing trajectories unique to that skill. Thus, this reward term is maximised when there is no uncertainty in the prediction.
  \item[($\beta$)] The probability model, $p_\rvtheta(\rvg \mid \rvs_0)$, parametrised by $\rvtheta$, represents the goal-selection policy. As part of the reward, the purpose of this term is to reinforce high entropy: the agent should keep pursuing a wide range of goals, for each starting state, $\rvs_0$, as the reward is higher when the goal had a low probability of being selected.
\end{description}

Given a starting state, $\rvs_0$, a goal, $\rvg$, is selected according to the distribution defined by $p_\rvtheta$, which the agent pursues until the final state $\rvs_f$. Then, $p_\rvtheta$ is reinforced based on how empowering the corresponding skill was (estimated using Eq.~\ref{eq:vicreward}). The final state becomes the new starting state, $\rvs_0 \leftarrow \rvs_f$, and the process is repeated. This can motivate the agent to ($\alpha$) follow skills that lead to distinct final states and ($\beta$) learn a wide range of skills for any given situation. Learning a state-conditional policy over skills fits the understanding of competence as finding suitable opportunities to use particular skills.

\subsubsection{That one uses a particular capacity in the course of one's action (C2.b)}
\phantomsection \label{sec:c2b}

While \textit{\hyperref[sec:c2a]{C2.a}} is about finding suitable opportunities to use particular skills, \textit{\hyperref[sec:c2b]{C2.b}} is more about an agent being motivated by observing that a skill is in use at all. The first term of the \gls{vic} reward function (Eq.~\ref{eq:vicreward}, $\alpha$) can be adapted to emphasise an agent's ability to distinguish that a particular skill is in use, by only conditioning on the current state. The reward for one such system, \gls{diayn} \citep{eysenbach2019diversity}, can be defined\footnote{See \citealp[Appendix~A, p.~12]{eysenbach2019diversity} for discussion on a constant term omitted here.} as:
\begin{align}\label{eq:diayn}
R_t(\rvs_{t+1}, \rvg) := \log q_\rvphi(\rvg \mid \rvs_{t+1}),
\end{align}
where $q_\rvphi$ is a discriminator parametrised by $\rvphi$ (refer back to the text following Eq.~\ref{eq:vicreward} for detail). An agent is rewarded for its ability to distinguish the current goal, $\rvg$; thus, the overall return is maximised when the agent can perfectly infer which of its skills is in use at any given time.

\Gls{diayn}, in further differentiation from \gls{vic}, ignores the state of the world in its choice of skill---instead, choosing skills uniformly at random, thus learning skills that are empowering on average over the observation space. Fixing the distribution over skills as uniform, although not an intuitive choice for a psychologically plausible model of goal selection (except maybe in cases such as the absence of prior information) is common in the \gls{cbim} literature. This approach ensures that all skills receive, in expectation, an equal amount of training signal to improve. \Gls{diayn} learns a diverse set of skills---running, walking, hopping, flipping, and gliding---across a wide array of situations \citep[p.~5]{eysenbach2019diversity}. It thus shows that recognising which skill is in use, that is, being able to distinguish the use of a particular capacity in the course of one's action, can drive skill learning.

Beyond \gls{diayn} and \gls{vic}, one can turn to other variational empowerment methods \citep[see][]{choi2021variational} for further examples of \glspl{cbim} fitting competence as skill use.

\subsection{Task performance (C3)}
\label{sec:taskperformance}

Next, we discuss computational models for two possible ways of understanding \emph{task performance}: observing that one performs well at an intended task (\textit{\hyperref[sec:c3a]{C3.a}}) and performing to an extent that requires a certain skill or skill level (\textit{\hyperref[sec:c3b]{C3.b}}).

\subsubsection{Observing that one performs well at an intended task (C3.a)}
\phantomsection \label{sec:c3a}

Given an agent has set itself a goal (task), if an agent has a mechanism to estimate its progress towards a chosen goal, it can use that estimate as a proxy to track its own performance on the task. An example system of this kind is \gls{rig} \citep{nair2018visual}. \Gls{rig} encodes the agent's observations into a lower-dimensional latent space from which the agent samples goals to pursue. Defining the reward as the negative distance between the goal and the current state in latent space encourages the agent to move its observations closer to the goal:
\begin{align}\label{eq:rig}
R_t(\rvs_{t+1}, \rvg) := -\left\lVert\phi(\rvs_{t+1})-\phi(\rvg)\right\rVert_\rmA,
\end{align}
where $\phi(\rvs_{t+1})$ is the learned state representation of $\rvs_{t+1}$, $\phi(\rvg)$ denotes the goal drawn from the latent space, and $\rmA$ is a matrix that can give more importance to some dimensions of the latent space over others \citep[p.~5]{nair2018visual}.

In contrast to \gls{diayn}, in which the agent learns to master a fixed number of skills, in \gls{rig}, the agent selects from an infinite number of goals (the space is continuous). This results in a sort of smoothness of the space that the agent can benefit from: if one goal is sampled near another that the agent has already learned to achieve, it may be able to generalise and use some parts of an existing policy to more efficiently learn to achieve the new goal. In the \gls{rig} system, \emph{goals} take on the intuitive meaning as desired states for the agent to reach (discussed in \textit{\nameref{sec:imgcrl}}), and ``performing well'' at an intended task is operationalised as minimising the semantic distance between the agent's observations and its self-generated goals. Over time, \gls{rig} agents learn to generate goals that approximate actual observations they make of their environment, while simultaneously learning the skills necessary to consistently perform well (reach the goals).

\subsubsection{Specifically to an extent that requires a certain skill or skill level (C3.b)}
\phantomsection \label{sec:c3b}

While \gls{rig} captures the idea of motivation by performance on a self-selected task (\textit{\hyperref[sec:c3a]{C3.a}}), it does not take into account the level of skill needed to perform well on the task (\textit{\hyperref[sec:c3b]{C3.b}}). For this, we can look to other system designs inspired by the idea of preferring to focus on goals or skills at an appropriate level of difficulty, given what the agent already knows. To identify skills of appropriate difficulty, such systems can make use of \gls{acl} (for a review see \citealp{portelas2021automatic}), enabling an agent to self-organise their goal selection. This can significantly reduce the number of training examples required for an agent to learn to reach its goals \citep[p.~1]{florensa18automatic}.

For example, in the \gls{curious} system by \citet{colas2019curious}, the system considers its capability to achieve its goals. Specifically, the agent is motivated to pursue goals in certain parts of the sensory space in which it believes itself to be increasingly or decreasingly successful at achieving its goals, estimated by measuring how successful it has been in the recent past---that is, its \gls{lp}. The underpinning logic of the system is that such goals will be of optimal difficulty, neither too easy nor too difficult. In the \gls{curious} experiments, an agent manipulates cubes by learning to control a robotic arm (\citealp[Supplementary p.~3]{colas2019curious}). Each observation includes information about the gripper and the objects of interest (e.g., their position and rotation), and each goal is a desired observation. The high-dimensional observation space makes learning challenging. Furthermore, some parts of the sensory space are irrelevant for achieving certain goals---for example, if the goal is to manipulate only one of several cubes, the agent might not need information about the other cubes. To ignore such irrelevant information in its formulation of goals, \gls{curious} modularises the observation space, which is also the goal space, with each module being a specific, hand-defined subspace of this space. A \gls{curious} agent estimates its own subjective \gls{lp} for each module, and uses the absolute values of its \gls{lp} estimates to compute a probability distribution for selecting the module to generate a goal within \citep[p.~4]{colas2019curious}. This mechanism helps the agent avoid modules containing goals that are too difficult or for which it already has sufficient skill, as well as to return to skills it is forgetting \citep[pp.~2--3]{colas2019curious}. In this way, the selection of different goals is driven by self-assessment of task performance.

The \emph{subjective \gls{lp}} for a given module, $LP(M_i)$, is defined as the derivative of \emph{subjective competence}, $C(M_i)$, with respect to time, where $C(M_i)$ is defined as the mean over results from self-evaluated trials performed by the agent (see \citealp[p.~4]{colas2019curious}). The results are stored in a queue of preset length, and each result is binary: either 1 for a successful trial or 0 for a failure. This can be viewed as the frequentist estimate of the probability of success. Finally, the probability of choosing a module is computed as:
\begin{align}\label{eq:curiouslpprob}
p(M_i) := \epsilon \times \frac{1}{N} + (1-\epsilon) \times \frac{|LP(M_{i})|}{\sum_{j=1}^N |LP(M_{j})|},
\end{align}
in which $\epsilon \in [0,1]$ is a hyperparameter that can be used to ensure that even poorly performing modules are occasionally selected, and $N$ corresponds to the total number of modules. As $\epsilon$ approaches 1, the probability distribution over modules resembles a uniform distribution. When $\epsilon$ is smaller than 1, the modules with the highest absolute \gls{lp} will have the highest probability of being selected. %

In the same vein, an emerging \gls{acl} paradigm, \gls{ued} (for a review see \citealp{rutherford2024no}), formalises the problem of automatically generating environments for agents based on some criterion, for instance, the difference between the current agent and (in practice, an approximation of) an optimal agent \citep[p.~1]{rutherford2024no}. In \gls{ued}, environments with their own reward functions can be thought of as goals for an agent to pursue. As part of one such system, \gls{omni-epic}, agents self-generate novel, interesting, and solvable goals, appropriate for an agent's current capabilities \citep[p.~2]{faldor2024omni-epic}.

While a significant proportion of \gls{acl} methods optimise task difficulty or competence progress with respect to current capabilities, \gls{acl} can also motivate agents to optimise goal selection based on other criteria that can be interpreted as proxies for ``skill level'' (from \textit{\hyperref[sec:c3b]{C3.b}}). \citet{pong2020skew-fit} extended \gls{rig} (introduced in \textit{\hyperref[sec:c3a]{C3.a}}) by weighting the selection of previously visited states used in learning the goal-generation model, encouraging the agent to generate goals resembling less frequently visited states. This weighting can be seen as a heuristic for attending to tasks that are expected to be non-trivial for the agent. As a second example, \citet{lintunen2024diversity} introduce a method motivating agents to select goals based on how much the agent believes pursuing them will diversify its entire set of skills. As \gls{acl} systems often consider the level of skill needed to perform well on the task, they offer a good starting point for formalising \textit{\hyperref[sec:c3b]{C3.b}}.

\subsection{Capacity growth (C4)}
\label{sec:capacitygrowth}

Competence, understood as \emph{capacity growth}, can be interpreted in two different ways: observing gain in the strength (\textit{C4.a}) and/or range (\textit{C4.b}) of one’s skills. While we saw that a reward reflecting a derivative of some measure of performance (e.g., \gls{lp}) can direct the agent towards goals requiring a certain level of skill (\textit{\hyperref[sec:c3b]{C3.b}}), the same derivative directly encourages capacity growth or improvement (\textit{C4.a}). While these two facets appear to have different meanings, from a computational perspective, they can be closely aligned.

An example of a system that explores both \textit{C4.a} and \textit{C4.b}---both repeatedly engaging in skills that need strengthening and extending the agent's skill repertoire---is \gls{imrl} described by \citet{barto2004intrinsically} and \citet{singh2004intrinsically}. \Gls{imrl} extends the agent's repertoire of skills (\textit{C4.b}): each time a ``salient'' event is observed, the agent adds a new skill associated with the salient event.\footnote{\citet{barto2004intrinsically} leave to future work what ``salient'' means and how to determine whether an event is salient or not. They simply hard-code events like turning on a light or ringing a bell as salient.} After this first observation, the agent learns how to return to observe the salient event again from any other situation via experience over time---this is the new skill associated with the salient event. Strictly speaking, the skill repertoire expands as a matter of course, not because the agent is intrinsically rewarded for expanding it.

The intrinsic reward used in the \gls{imrl} system \citep[p.~4]{singh2004intrinsically} can be defined as:
\begin{align}
    R_t(\rvs_t,\rvs_{t+1},\rvg_{\rvs_{t+1}}) :\propto \left\{ \begin{array}{ll}
        1 - P^{\rvg_{\rvs_{t+1}}}(\rvs_{t+1} \mid \rvs_t ) & \text{if }\rvs_{t+1}\text{ is salient}\\
        0 & \text{otherwise}
        \end{array}\right.
\end{align}
where $P^{\rvg_{\rvs_{t+1}}}$ is what \citet{precup1997multi} call a \emph{multi-time model} for skill $\rvg_{\rvs_{t+1}}$. $P^{\rvg_{\rvs_{t+1}}}$ is not \textit{quite} a probability estimate for the transition from $\rvs_t$ to $\rvs_{t+1}$ given the agent is following skill $\rvg_{\rvs_{t+1}}$, but that is close to the correct intuition. %

The agent only receives a non-zero intrinsic reward when it reaches a salient state. The intrinsic reward here can be thought of as modelling how ``surprising'' \citep[p.~6]{singh2004intrinsically} it is that the agent reached $\rvs_{t+1}$ immediately from state $\rvs_t$---which it has just done---by following the skill $\rvg_{\rvs_{t+1}}$. If the agent expected, without uncertainty, to transition directly from $\rvs_t$ to $\rvs_{t+1}$, then $P^{\rvg_{\rvs_{t+1}}}$ is 1 and the reward is 0 (the transition is very ``unsurprising''). If the agent did not expect it could transition from $\rvs_t$ to $\rvs_{t+1}$ by following skill $\rvg_{\rvs_{t+1}}$, then $P^{\rvg_{\rvs_{t+1}}}$ is 0 and the reward is 1 (maximally ``surprising''). If the agent is uncertain, $P^{\rvg_{\rvs_{t+1}}}$ will be somewhere in between. However, if the agent expected that it could reach $\rvs_{t+1}$ from $\rvs_t$ by following skill $\rvg_{\rvs_{t+1}}$, just that it would take more than a single step, $P^{\rvg_{\rvs_{t+1}}}$ will be discounted based on the length of the path.
This reward encourages the agent to decrease the number of steps it takes to get from one state to another, strengthening its skill of reaching a particular state.

Another example of \textit{C4.b}, that is, being motivated by growing the number of skills in the agent's repertoire, is the aforementioned \gls{vic} (see \textit{\hyperref[sec:c2a]{C2.a}}). The \gls{vic} reward function motivates the agent to grow its number of effective skills: in expectation, the second term of the reward function (Eq.~\ref{eq:vicreward}, $\beta$) increases as the distribution over skills becomes more uniform. In theory, this encourages the distribution over skills not to collapse to a small number of skills,\footnote{\citet[Appendix E, pp.~17--18]{eysenbach2019diversity} showed empirically that, in practice, since there are numerous other factors that affect how agents learn, this collapse may still happen.} maintaining or increasing the size of the repertoire.

\section{Advancing Competence Modelling in SDT}
\label{sec:advancing-SDT}

What is gained for human motivation researchers by identifying computational formalisms that match aspects of verbal theory? For one, translating a concept like competence into an algorithmic description induces a different way of thinking about theory: the process forces us to specify abstract and informal verbal theories to the point that they can be turned into a system that can actually produce observable behaviour. Observing how a computational model of a theory behaves can ``reveal implicit assumptions and hidden complexities'' \citep[p.~4]{marsella2010computational} and resolve jingle-jangle fallacies (\citealp[pp.~3--4, 6]{murayama2025critique}; \citealp[p.~26]{pekrun2024overcoming}).

For instance, \gls{lp}-based methods can simultaneously exemplify skill use as (\textit{\hyperref[sec:c2a]{C2.a}}) an opportunity to use a particular skill, (\textit{\hyperref[sec:c3b]{C3.b}}) task performance as requiring a certain skill or skill level, and (\textit{\hyperref[sec:capacitygrowth]{C4.a}}) capacity growth as observing gains in the strength of one's skills. Given that \gls{sdt} does not acknowledge distinct facets, finding such a formalism that fits multiple facets supports the theory. On the other hand, in our matching, some facets are formally distinct. For example, it is unclear how (\textit{\hyperref[sec:effectance]{C1}}) effectance, (\textit{\hyperref[sec:capacitygrowth]{C4.b}}) being motivated by extending one's repertoire of skills, or (\textit{\hyperref[sec:c3a]{C3.a}}) being motivated by observing that one performs well at an intended task, equate to each other and other facets. By identifying formalisms for each of the facets, and testing whether they result in demonstrably different observable behaviours (e.g., using simulation), we can prompt review of \gls{sdt}---is one term, ``competence'', being used for multiple different constructs?

Another benefit of this matching is that it reveals underlying preconditions that \gls{sdt} fails to specify, such as that a competence-motivated agent must be able to recognise when its actions cause effects (\textit{\hyperref[sec:effectance]{C1}}) and differentiate between distinct skills (\textit{\hyperref[sec:skilluse]{C2}}, \textit{\hyperref[sec:taskperformance]{C3}}, \textit{\hyperref[sec:capacitygrowth]{C4}}). Similarly, formalisms of concepts such as novelty, diversity, and progress towards goals, underlie competence-driven learning in computational \gls{im}, yet are often overlooked in \gls{sdt}. By drawing on such formalisms from computational agents that demonstrably and autonomously develop capacities in diverse environments, this paper shines light on overlooked preconditions and what potential mechanisms might computationally fulfil them.

We do not argue that identifying AI algorithms that demonstrably perform some capacities will mean that we have identified which algorithms best approximate human cognition \citep[p.~626]{van2024reclaiming}. Rather, our work can support a cycle of theory development by inspiring new computational models, which can then be tested to refine the theory.

\section{Future Work and Conclusion}
\label{sec:futureconclusion}

Showing that \gls{rl} formalisms can be aligned with \gls{sdt} uncovers a wealth of resources for motivation research, with existing implementations of models, simulators, and evaluation metrics in the papers associated with each formalism. These resources can, by enabling large-scale simulation studies, allow us to explore computational \glspl{im} as candidates for model fitting to human data. We have used the different facets of competence outlined by \citet{deterding2024why} as starting points for selecting promising computational formalisms. However, it remains an open question whether the facets capture different cognitive mechanisms or even require separate formalisms to generate associated explananda, such as exploratory and challenge-seeking behaviour. Towards investigating part of this question, the computational \gls{rl} literature offers different methods for composing reward functions \citep[e.g.,][]{bagot2020learning}, but how best to combine rewards and what methods would be psychologically plausible still require work. Such methods, in combination with the aforementioned resources, allow us to study how distinct facets of competence, either individually or in conjunction, contribute to competence-driven behaviour and motivation more widely.

These avenues of research are contingent on developing effective means to compare artificial agents with biological agents (a project partially undertaken by researchers in \emph{representational alignment}; e.g., \citealp{sucholutsky2024getting}). While work exists comparing behaviour induced by different intrinsic reward functions (e.g., \citealp{santucci2012intrinsic,biehl2018expanding,linke2020adapting}), as does work studying different choices of state representation (e.g., \citealp{burda2019large}), more research is needed on how different models behave. It is also an open question how to best evaluate intrinsically motivated open-ended learning \citep[pp.~1167--1169]{colas2022autotelic}.

Most formalisms we identified relate to pursuing goals, so we need to understand how humans represent and generate goals \citep[p.~1]{colas2024what}. The topic has not been well studied for various reasons, e.g., computational intractability \citep[p.~1]{byers2024modeling}. Excitingly, goals have recently emerged as a multidisciplinary area of research spanning \gls{ai}, cognitive science, and developmental psychology (e.g., \citealp{colas2022autotelic,chu2024in,davidson2024goals}). \citet[p.~1150]{molinaro2023a} suggest that goals affect many agent dynamics, such as representing the environment, choosing relevant actions, and evaluating rewards.

In summary, we demonstrate that computational models of \gls{im} can plausibly match verbal definitions of competence in \gls{sdt}.
Specifically, we show that four distinct facets of competence---effectance, skill use, task performance, and capacity growth---can be aligned with existing formalisms from intrinsically motivated \gls{rl}, and in the process, highlight the potential of computational models to provide deeper insights into motivated behaviour. While this is a promising first step, further research is required to empirically study these computational models and consider other potential models, inviting collaboration from motivation researchers across disciplines.

\section{Acknowledgments}

We thank the members of the Intrinsically Motivation Open-ended Learning (IMOL) community for fun and illuminating discussions during workshops in Paris (2023) and Vancouver (2024). We also thank the members of the Autotelic Interaction Research (AIR) group, in particular Niki Pennanen, for feedback. EML and CG received financial support from the Research Council of Finland (NEXT-IM, grant 349036) and NMA from the Helsinki Institute for Information Technology.

\bibliographystyle{apalike}
\setlength{\bibleftmargin}{.125in}
\setlength{\bibindent}{-\bibleftmargin}
\bibliography{refs}

\end{document}